# Wide and deep volumetric residual networks for volumetric image classification

Varun Arvind [1], Anthony Costa [2], Marcus Badgeley[2], Samuel Cho[1], Eric Oermann [2]

[1] Department of Orthopedics, Icahn School of Medicine at Mount Sinai, 1 Gustave Levy Pl. New York, NY 10029
[2] Department of Neurosurgery, Icahn School of Medicine at Mount Sinai, 1 Gustave Levy Pl. New York, NY 10029

**Abstract**
3D shape models that directly classify objects from 3D information have become more widely implementable. Current state of the art models rely on deep convolutional and inception models that are resource intensive. Residual neural networks have been demonstrated to be easier to optimize and do not suffer from vanishing/exploding gradients observed in deep networks. Here we implement a residual neural network for 3D object classification of the 3D Princeton ModelNet dataset. Further, we show that widening network layers dramatically improves accuracy in shallow residual nets, and residual neural networks perform comparable to state-of-the-art 3D shape net models, and we show that widening network layers improves classification accuracy. We provide extensive training and architecture parameters providing a better understanding of available network architectures for use in 3D object classification.

## 1. Introduction

Multi-dimensional computer vision problems are becoming increasingly common with the emergence of light detection and ranging (LiDAR), medical imaging, computer assisted design (CAD), and virtual reality datasets. Performing deep learning in these 3D, and in some cases 4D, data spaces introduces novel computational challenges due to the increase in dimensionality of the underlying problem as well as the challenge of engineering larger deep learning models. The limitations of available compute resources place a constraint on architecture choice, and seems to limit the feasibility of extending the current generation of extremely deep convolutional neural networks (CNNs) to 3D applications. We present our results with developing volumetric CNN architectures that utilize wide residual layers on classification on the ModelNet-40 dataset.

## 2. Methods

### 2.1 Data Set

The ModelNet-40 dataset consists of 12,311 3D CAD models from 40 classes with 9,843 training and 2,468 testing objects (Figure 1). These classes include common household objects, vehicles, and other items saved as vectorized CAD drawings. The CAD objects can be converted to voxels of an arbitrary dimension, with generally accepted dimensions for classification tasks being (30,30,30). There is no color channel like in many 2D imaging cases, the voxels are simply present or absent.

### 2.2 Model architectures and training

**Overview**
We implement residual neural networks and examine the effect of residual layer width on volumetric object classification.

**Network Framework**
Our residual neural network architecture is visualized in Fig 1, where the basic convolutional networks are as initially described by Zagoruyko et al.[1] Identity mappings in residual networks can be represented by the following:

$$y_t = h(x_t) + F(x_t, W_t)x_{t+1} = f(y_t), h(x_t)$$

$$= x_t \,, f(x) = \begin{cases} 0 \text{ for } x < 0 \\ x \text{ for } x \geq 0 \end{cases}$$

Where $x_{t+1}$ and $x_t$ are input and output of the t'th layer in the network. Fis a residual function and $W_t$ are its associated parameters. Here we employ a dropout layer into each residual block as a means of regularization. The general structure of out network is outlined in Table 1: an initial convolution layer (Conv3D) followed by max pooling, a series of residual blocks, average pooling and a final classification layer. Batch normalization and activation preceding 3D convolution was performed in all residual blocks.[2] A kernel size of 3 x 3 x 3 was used for all convolutions. Further we define widening factor k, where k multiplies the number of features in convolutional layers. This was used to evaluate the effect of further widening of the residual layers on classification accuracy (Table 1).

Table 1: We experimented with several widening factors, and noted that increasing model width markedly increased the number of trainable parameters. The maximum validation accuracy was reached with an intermediate widening factor (k=8).

| Widening Factor | Parameters | Accuracy |
|---|---|---|
| 1 | 122,032 | 0.7396 |
| 2 | 341,688 | 0.7768 |
| 4 | 1,081,672 | 0.7817 |
| **8** | **3,764,328** | **0.7949** |
| 16 | 14,826,408 | 0.7797 |

**Data Augmentation**

3D object classification is highly sensitive to rotational orientation. Training samples were augmented by randomly rotating each 3D shape along a randomly designated axis to increase inter-batch variance and model generalizability (Figure 2). The random rotations encourage the model to learn rotationally invariant features improving classification independence to rotational variation. The natural dataset contains objects in a variety of rotations, and therefore augmenting the dataset by employing random, single axis, 360 degree rotations is a reasonable means of data augmentation.

**Training**

Our model was implemented in the Keras Python module with the TensorFlow backend. Training was parallelized on a GPU cluster consisting of two consumer GTX1080s (nVidia) graphics cards accounting for a total of 5,120 CUDA Cores with 8Gb of memory per GTX1080.

Training data was randomly shuffled and partitioned into batch sizes of 64 to train within resource allocation constraints. Network training was achieved by using stochastic gradient descent with Nesterov momentum. Optimizer parameters used are as follows $\boldsymbol{\beta}1$of 0.9, $\boldsymbol{\beta}2$ of 0.999, $\boldsymbol{\varepsilon}$ of 1e-08, and a schedule decay constant of 0.04. We applied a learning rate annealing scheduler that dropped the learning rate from 0.0002 by a factor of .02 following plateauing of validation loss.

**Ensemblings**

Ensembling of our models was performed to improve generalizability and accuracy of volumetric classification. It is well known that ensembling of models consistently outperforms a single best model for classification tasks; however, constructing such ensembles is tedious. As reported by Huang et al., sequential single-model ensembling, snapshot ensembling, is an attractive tool that to address this problem.[3] During training, weights were saved every epoch for 10 epochs and a final model consisting of an ensemble of weights from each epoch was used. To compare, a single best model was trained 10 times, each for 10 epochs, and a final model was ensembled from the 10 independently trained models.

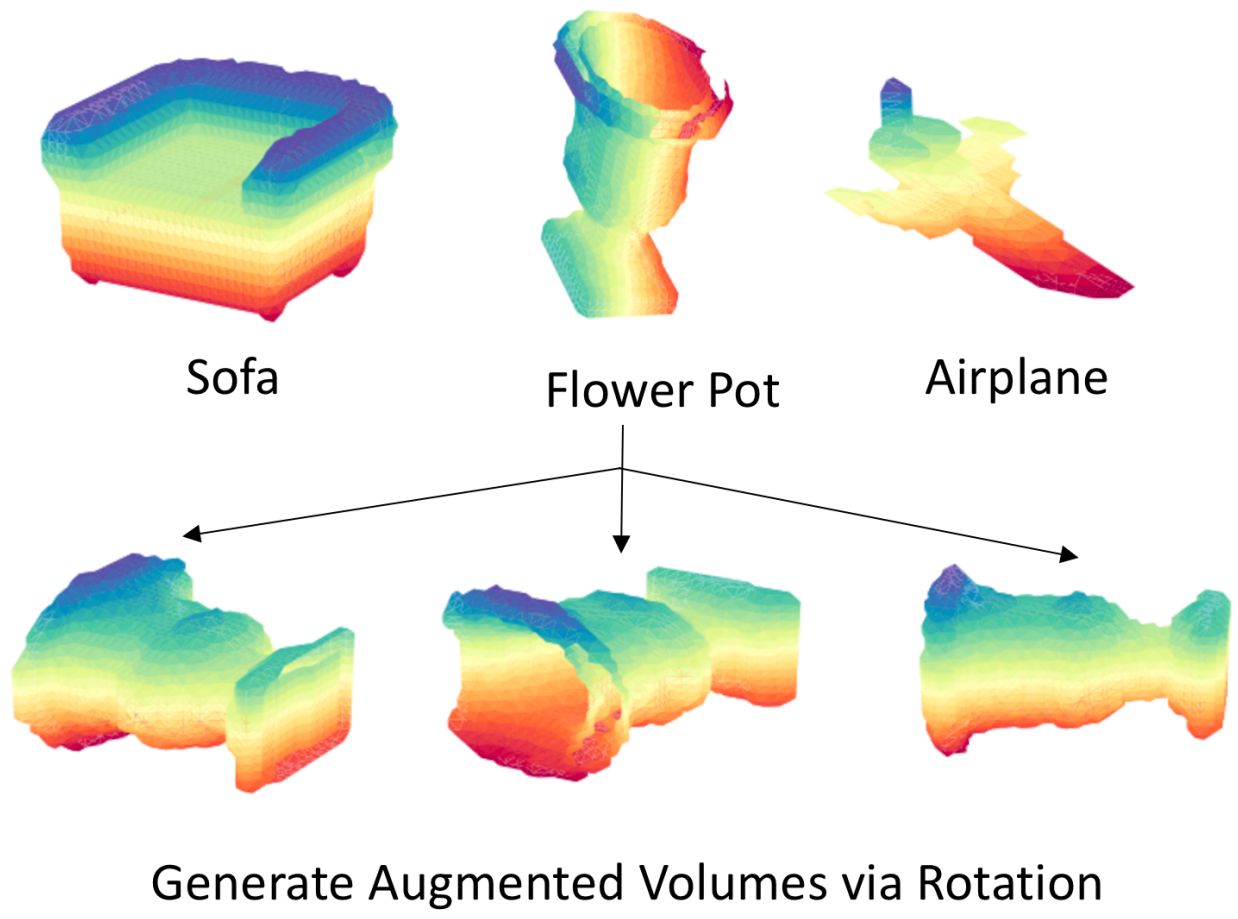

Figure 1: a. Residual blocks form the basis of most modern CNN architectures and utilize skip connections to better propagate information forwards and backwards through the network. b. The basic architecture of a residual network relies on residual block

## 3. Theory

CNNs are a type of deep learning algorithm originally designed for computer vision tasks.[4–6] By utilizing multiple layers of spatial convolutions, CNNs can learn hierarchical feature maps from the input data space to some output label space to obtain state-of-the-art performance across many standard computer vision tasks.[6–8] These computer vision tasks have been facilitated by the ready availability of two-dimensional imaging as well as tools for labelling 2D images such as MNIST and ImageNet.[4,88,9] There has been substantial work in applying and refining 2D-CNNs to these datasets, which have resulted in numerous architectural improvements, most recently the use of skip connections and residual blocks.[7,10]

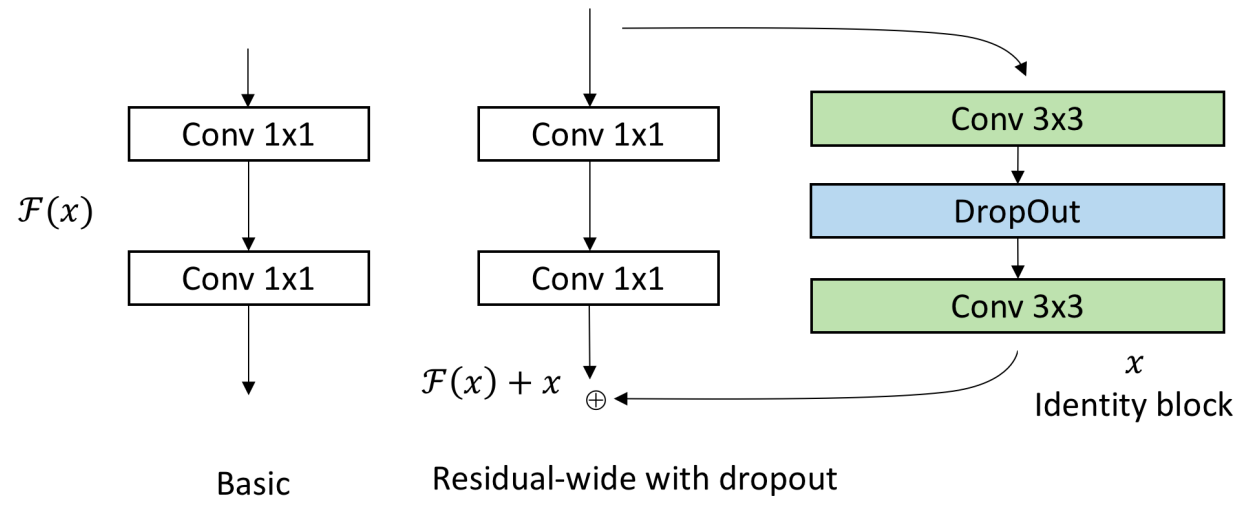

| Group Name | Output Size | Block Type |
|---|---|---|
| Conv3D | 30 x 30 x 30 | [3 x 3 x 3, 8 x k] |
| ConvBlock3D | 8 x 8 x 8 | [3 x 3 x 3, 8 x k] |
| IdentityBlock3D | 8 x 8 x 8 | [3 x 3 x 3, 8 x k] |
| IdentityBlock3D | 8 x 8 x 8 | [3 x 3 x 3, 8 x k] |
| ConvBlock3D | 8 x 8 x 8 | [3 x 3 x 3, 16 x k] |
| IdentityBlock3D | 8 x 8 x 8 | [3 x 3 x 3, 16 x k] |
| IdentityBlock3D | 8 x 8 x 8 | [3 x 3 x 3, 16 x k] |

Figure 2: a. The ModelNet40 dataset consists of several CAD objects from general classes including sofas, flower pots, and airplanes. b. These volumes can be augmented using standard image data augmentation techniques such as rotation extended to three dimension

Skip connections and residual blocks (Figure 1a) facilitate the direct flow of information throughout the length of a network for both forwards and backwards passes, such that intermediate features are comprised of the sum of preceding residual layers rather than the product (Figure 1b).[2] This enables particularly deep networks to learn highly sophisticated feature maps.[2] In addition to their depth, residual networks can benefit from variable width, with the major limitation being number of trained parameters.[1]

The extension of CNNs to the 3D domain has been somewhat gradual due to the absence of large, canonical datasets relative to 2D dataset. The ModelNet-40 dataset directly addresses this absence of high quality, standardized classification datasets by providing 40 classes of CAD objects.[11] There have been numerous classifications efforts surrounding the ModelNet-40, looking at both 2D CNNs that utilize projections of the objects as well as true 3D-CNNs.[11–14] Performing convolutions in 3D dimensions is a natural approach to the problem, at the expense of a significant increase in free parameters and computational demand.

## 4. Results

We experimented with several widening factors and noted that even small increments of the widening factor significantly increased in the number of trainable parameters (Table 1). Furthermore, increases in widening factor dramatically improve model performance during training, however too much widening of residual layers lead to a loss in validation accuracy (Figure 3: a-b). Following a grid-search optimization of training hyper-parameters, our model achieved an accuracy of 82.03% (Figure 4).

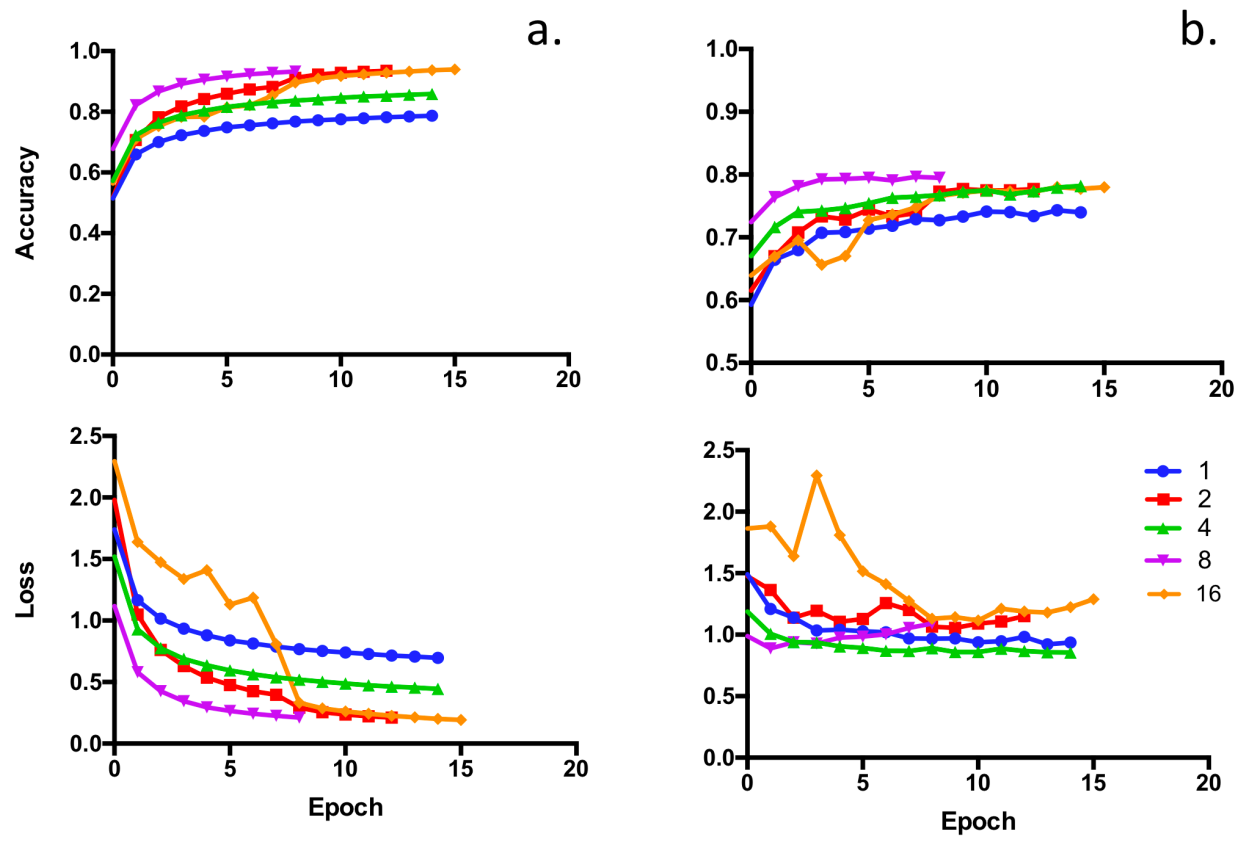

Figure 3: Shown are a. training and b. testing accuracy and loss curves for residual nets with widening factor (k) values of 1, 2, 4, 8, and 16.

Our best result came with ensembling (Table 2). We built two separate ensembles, one of 10 independently trained models, and a second of an ensemble of a single model with weights saved at epochs sequentially along in its training process. The independent model ensemble took approximately 108 hours to train and achieved a final validation accuracy of 0.8650, while the sequentially ensembled model took 10 hours to train and achieved a final validation accuracy of 0.8537. Notably ensembling the independent models lead to an increase in performance of 4.47%, while ensembling a single best model sequentially across epochs lead to a relative gain in validation accuracy of 3.34% relative to the optimized model (Table 2).

Table 2: We attempted two different types of ensembling, one involved multiple independent models while the other utilized a weights from a single model saved at sequential epochs during training as motivated by Huang et al.[3]

| | **Accuracy** |
|---|---|
| Unoptimized | 0.7949 |
| Optimized | 0.8203 |
| Ensembling | **0.8650** |
| Huang Ensembling | 0.8537 |

We plotted a confusion matrix of the various object categories (Figure 5). Notably several categories had very few examples in the validation dataset (cup 16, flower pot 20 training examples). Expectedly, our best performing model had the greatest difficulty with classifying similar objects, such as dressers and nightstands. 13.7% of dressers were classified as night stands and 26.2% of night stands classified as dressers.

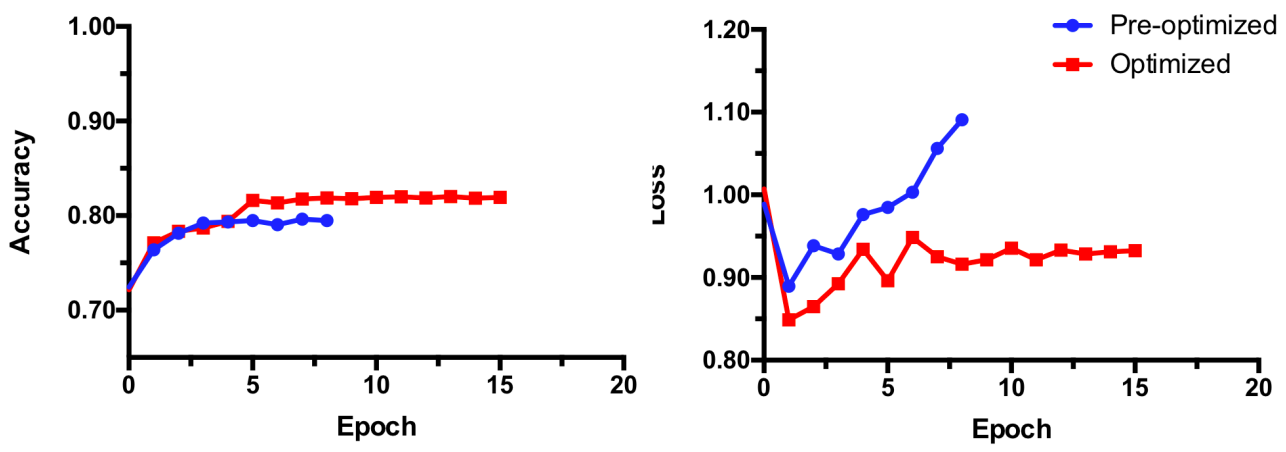

Figure 4: Optimization of residual net width. A grid search was performed to produce an optimized wide residual net (red) and compared to the pre-optimized neural net (blue).

## 5. Discussion

Our results for classifying CAD objects with volumetric residual network architectures approach state-of-the-art results on the ModelNet-40 leaderboard (http://modelnet.cs.princeton.edu/) (Table 3). Echoing the findings of Zagoruyko and Komodakis, we found that widening the residual network architecture was beneficial to a point.[1] As our investigation of the problem space shows, several objects are underrepresented and likely require more original data or more extensive augmentation to learn better classifiers.

We also note that the features for learning these objects are quite constrained due to the volumetric dimensions. Common 2D datasets have much higher resolutions (e.g., ImageNet (256x256). Importantly, the overall low resolution (30,30,30) of the volumetric data space makes this for a particularly challenging task. Distinguishing such objects at (30,30,30) is challenging for humans, and unsurprisingly challenging for our models. Studies in two dimensional images have shown that even small decreases in resolution can lead to a substantial loss of features that are essential for classification tasks.[15,16] We suspect that higher resolution inputs would result in significantly improved classification results. Despite the low resolution though the underlying data space is still quite large due to the volumetric nature of the problem and the convolutional operation must be carried across an additional spatial dimension. Secondly, as expected, the classes with the greatest error are objects that naturally appear quite similar. Misclassifications were particularly frequent for dresser (class #14) and nightstand (class #23), where overall appearance is similar and distinguishing features are obscured by the low resolution.

There were notable challenges in fitting the volumetric models into the memory of our graphics cards. We generally were forced to employ small mini-batch sizes of 64, which resulted in longer training times to convergence. A further computational refinement which we did not employ, but suspect would have yielded significant performance benefits, was quantization of our models from single precision (fp32) to int8. Compression in model space via quantization has been demonstrated to result in substantially smaller models, over 10x compression, and to be lossless for image classification models such as Alexnet.[17]

We attempted two different strategies for generating the ensembles. A traditional approach of training N independent models resulted in slightly improved results relative to training a single model and ensembling saved copies of it at N different training epochs. However, the latter approach had a substantially shorter train time and consumed significantly less computational resources while still yielding most of the benefits of ensembling.

Given the limitations associated with the ModelNet-40 dataset, our current work presents a highly promising direction forwards with volumetric residual networks employing variable width and depth. Furthermore, creative ensembling can obtain near-optimal results while markedly reducing computational burden and training time.

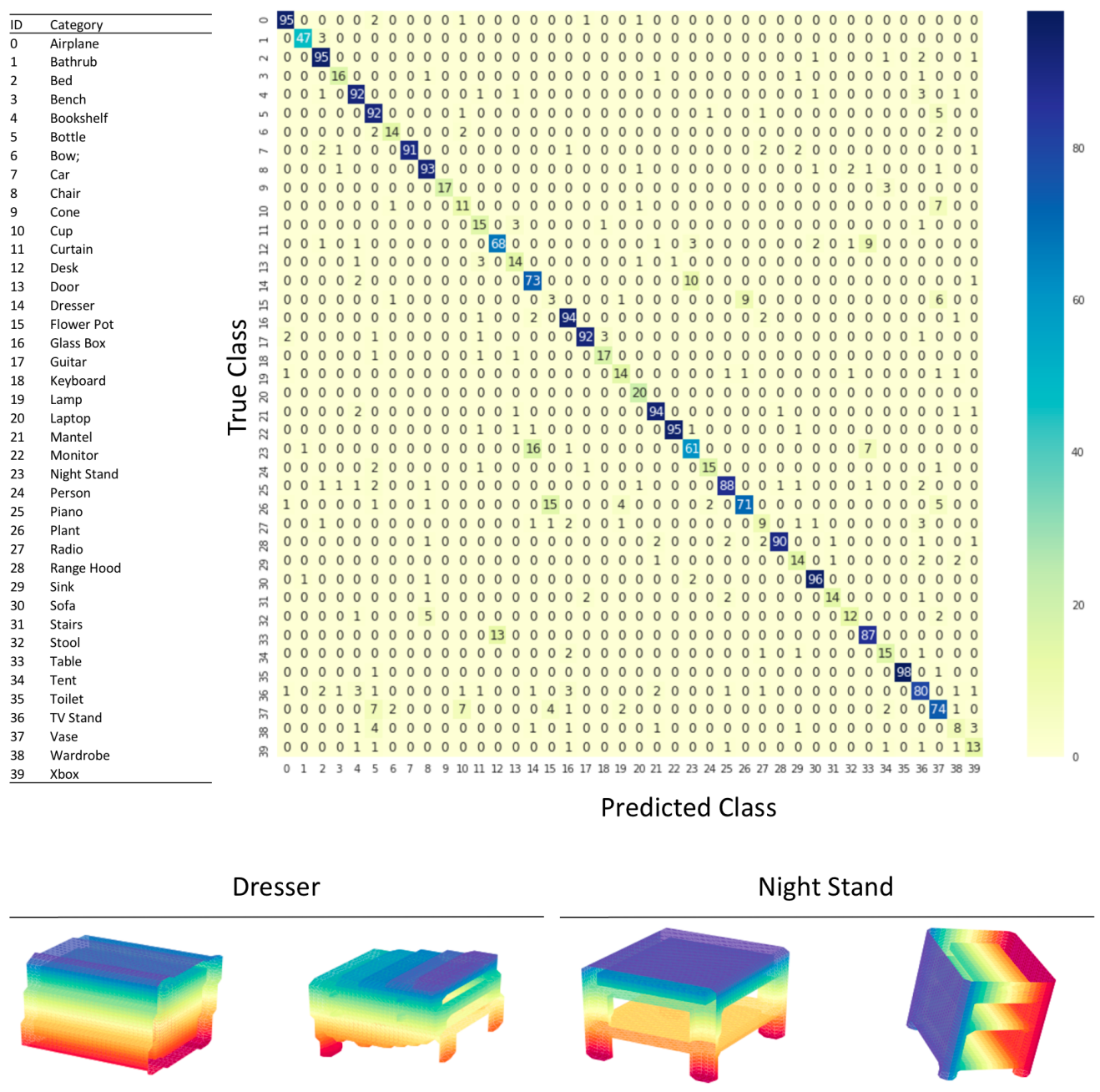

Figure 5: Confusion matrix of the best performing model's outputs, and visualizations comparing highly misclassified categories. There are several categories which are poorly represented in the validation set. The model has difficulty with classifying similar objects (e.g. dresser(14) and nightstand(23)). Performance is likely limited by the overall low resolution (30,30,30) of the volumetric data space.

## 6. Conclusions

Wide, volumetric residual networks obtain state-of-the-art classification results on the volumetric dataset, ModelNet-40. Ensemble classifiers built from multiple epochs of a single network maximize classifier accuracy while preserving compute resources and saving training time. Further work needs to be performed on volumetric classification on higher resolution problems to fully investigate ways of optimizing volumetric CNNs for deployment on real world problems in medical imaging, autonomous driving, and virtual reality.